\def\year{2019}\relax
\documentclass[letterpaper]{article} 
\usepackage{aaai19}  
\usepackage{times}  
\usepackage{helvet}  
\usepackage{courier}  
\usepackage{url}  
\usepackage{graphicx}  

\usepackage{amsmath}
\usepackage{amsfonts}
\usepackage{subfig,booktabs,multirow}

\begin{document}
%
\title{RAP-Net: Recurrent Attention Pooling Networks \\for Dialogue Response Selection}
\author{Chao-Wei Huang\thanks{Huang and Chiang have equal contributions.}\quad Ting-Rui Chiang\footnotemark[1]\quad Shang-Yu Su\quad Yun-Nung Chen \\
National Taiwan University, Taiwan \\
\texttt{\{r07922069,r07922052,f05921117\}@csie.ntu.edu.tw\quad  y.v.chen@ieee.org}}

\maketitle
\begin{abstract}
The response selection has been an emerging research topic due to the growing interest in dialogue modeling, where the goal of the task is to select an appropriate response for continuing dialogues.
To further push the end-to-end dialogue model toward real-world scenarios, the seventh Dialog System Technology Challenge (DSTC7) proposed a challenge track based on real chatlog datasets.
The competition focuses on dialogue modeling with several advanced characteristics: (1) natural language diversity, (2) capability of precisely selecting a proper response from a large set of candidates or the scenario without any correct answer, and (3) knowledge grounding.
This paper introduces \emph{recurrent attention pooling networks} (RAP-Net), a novel framework for response selection, which can well estimate the relevance between the dialogue contexts and the candidates.
The proposed RAP-Net is shown to be effective and can be generalize across different datasets and settings in the DSTC7 experiments.

\end{abstract}

\section{Introduction}

With the increasing trend about dialogue modeling, response selection and generation have been widely studied in the NLP community.
In order to further evaluate the current capability of the machine learning models, a benchmark dataset was proposed in the seventh Dialog System Technology Challenge (DSTC7) \cite{DSTC7}, where the task is to select the most probable response given a partial conversation.
To approximate the real world scenarios, several variants of selections are investigated in this task: 1) selecting from 100 candidates, 2) selecting from 120,000 candidates, 3) selecting multiple answers, 4) there may be no answer, and 5) with external information.
Some subtasks are much more difficult than the original setting.
In addition, the ability of generalization should be examined; hence, two datasets, Ubuntu IRC dialogs \cite{kummerfeld2018analyzing} and course advising corpus, are utilized for the experiments.
These datasets have very different properties, where the dialogs in the Ubuntu IRC dialogs dataset are very technical, and are more problem-solving-oriented, while the dialogs in the course advising dataset tend to be more casual, and the goals are more like inquiring information rather than solving a specific problem.
In sum, the challenge covers a wide range of scenarios in real-world applications and serves as a set of benchmark experiments for evaluating dialogue response selection models.

Recently, deep neural networks have been widely adopted for end-to-end response selection modeling. 
The prior work generally employed two encoders to map the conversation and the response into vector representations, and then designed a classifier to measure the relation between these two representations.
An intuitive design is to encode two sequences separately via recurrent neural networks (RNNs) and then compute a score between the last hidden state of two RNNs~\cite{feng2015applying,mueller2016siamese,lowe2015ubuntu}.
The MV-LSTM \cite{wan2015deep} improved the design by deriving a similarity matrix between outputs of RNNs, and then used max-pooling and multi-layer perceptron (MLPs) to aggregate the similarity scores.
To better utilize the interactive information, other approaches employed the attention mechanism \cite{bahdanau2014neural} to facilitate the encoding process~\cite{tan2015lstm,rocktaschel2016reasoning,wang2016inner,santos2016attentive,shen2017inter,tay2018multi}.

Motivated by the prior work that effectively utilized attention mechanisms in diverse ways, this paper proposes a novel framework for dialogue response selection, called \emph{recurrent attention pooling networks} (RAP-Net).
The proposed model consists of (1) multi-cast attention network (MCAN) \cite{tay2018multi} for extracting features from input words, (2) feature-fusion layer for integrating domain-specific knowledge-grounded features and information from the MCAN layer, and (3) a proposed dynamic pooling recurrent layer for  extracting sentence-level information by pooling dynamically based on utterance boundaries. 
The proposed model is shown to be effective for different datasets (Ubuntu, Advising) and different settings (subtask 1, 3, 4) in the DSTC7 experiments.
Furthermore, the framework can also generalize to other retrieval-based tasks.

\section{Task Description}

In the response selection challenge, given a partial conversation and a set of response candidates, the system is required to select one response from the candidates set. 
A partial conversation consists of $l$ utterances: $U: \{u_1, u_2, \cdots, u_l\}$, an utterance is a sequence of words.
Each speaker participated in the conversation is given a special identifier, say \texttt{<speaker1>}, \texttt{<speaker2>}, and the special identifier is prepended to the utterances which speaker speaks.
So the $i$th utterance is denoted as  $u_i: \{w_{i,0}^U, w_{i,1}^U, w_{i,2}^U, \cdots, w_{i,n_i}^U \}$. A candidate set consists of $k$ candidates $X: \{x_1, x_2, \cdots, x_k \}$. And each candidate is a sequence of words $x_j: \{ w_{j,1}^X, w_{j,2}^X, \cdots, w_{j,m_j}^X \}$.
For some datasets, some word features grounded to specific domain knowledge are also available. The knowledge-grounded features of a word $w$ are denote as $F(w)$. 
Among the candidates, there would be either some correct responses or none. 
The labels indicating if the candidate are correct answers are denoted as $Y: \{y_1, y_2, \cdots, y_k\}$.

\section{RAP-Net: Recurrent Attention Pooling Networks}

\begin{figure}[t!]
  \centering
  \includegraphics[width=1\linewidth]{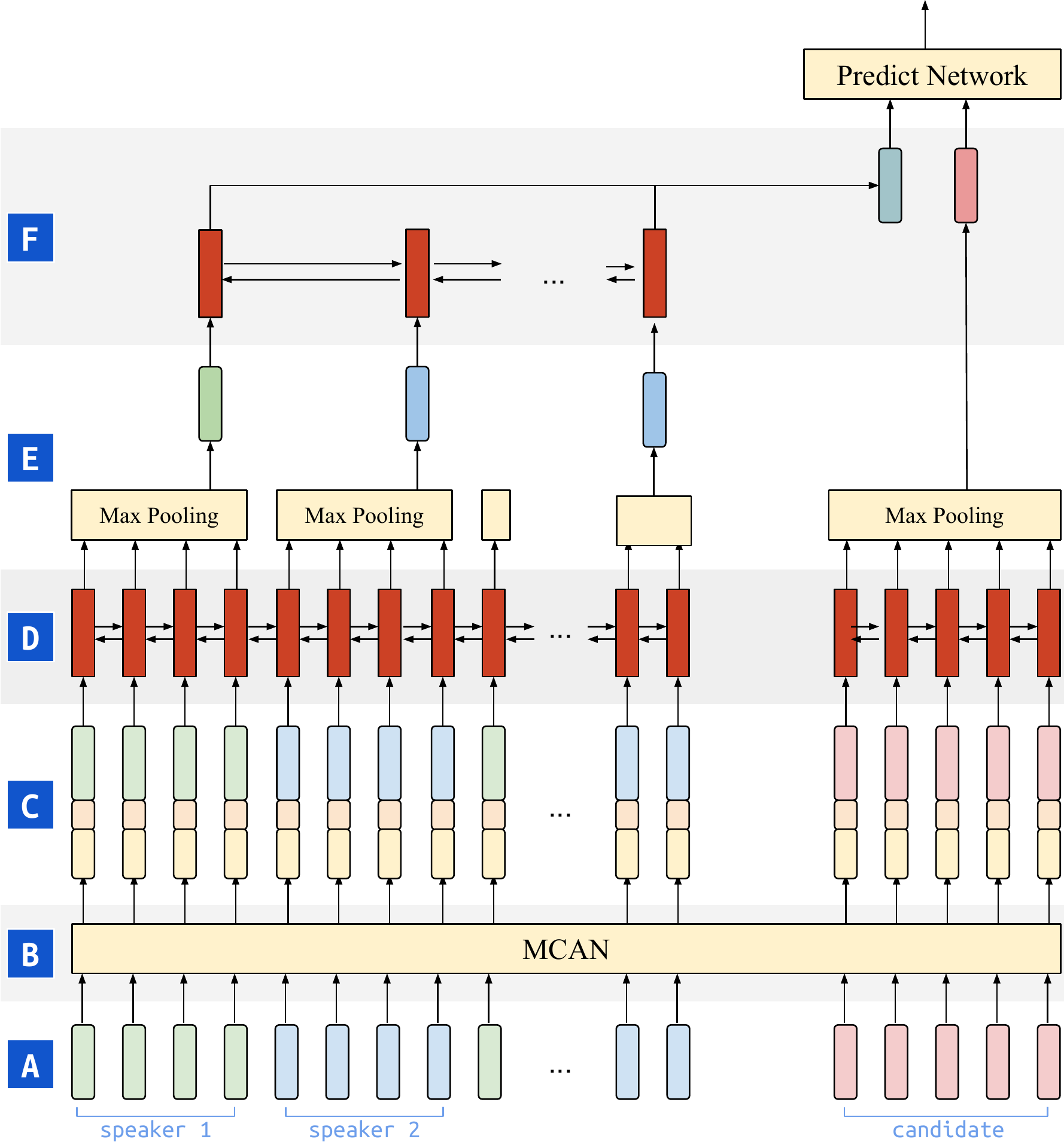}
  \caption{The architecture of the whole proposed model. (A) is the word embedding of utterances (light green and light blue color denotes words spoken by different speakers) and candidate (light red color). (B) is the MCAN. (C) Word embeddings along with extra features, which include knowledge grounded features (light orange color) and features extracted by the MCAN (light yellow color). (D) is the first bi-directional LSTM layer. (E) is the dynamically pooling layer. Note that LSTM outputs are grouped according to utterances. (F) is the second bi-directional LSTM layer. }
  \label{fig:solving}
\end{figure}

In this paper, we propose a novel framework for dialogue response selection illustrated in Figure~\ref{fig:solving}, and the four-step pipeline is described as follows. 

\subsection{Multi-Cast Attention Network (MCAN)}

First, the multi-cast attention network (MCAN)~\cite{tay2018multi} is applied to extract extra word features with various attention mechanisms on two word sequences.
Here we concatenate context utterances $d = [u_1, u_2, \cdots, u_l]$ as the first sequence, while $q = x_j$ as the sequence for the $j$th candidate.
For each word $w$ in either $d$ or $q$, we pass the word representation into a highway layer $H$ \cite{srivastava2015highway} to obtain a new representation $w'$:
\begin{align}
   & w' = H(w) = \sigma(W_g w) \odot \mathrm{ReLU}(W_h w)  + (1 - \sigma(W_g w))\odot w. \nonumber
\end{align}
where $W_g, W_h$ are parameters to learn.

\begin{figure}[t!]
  \centering
  \includegraphics[width=1\linewidth]{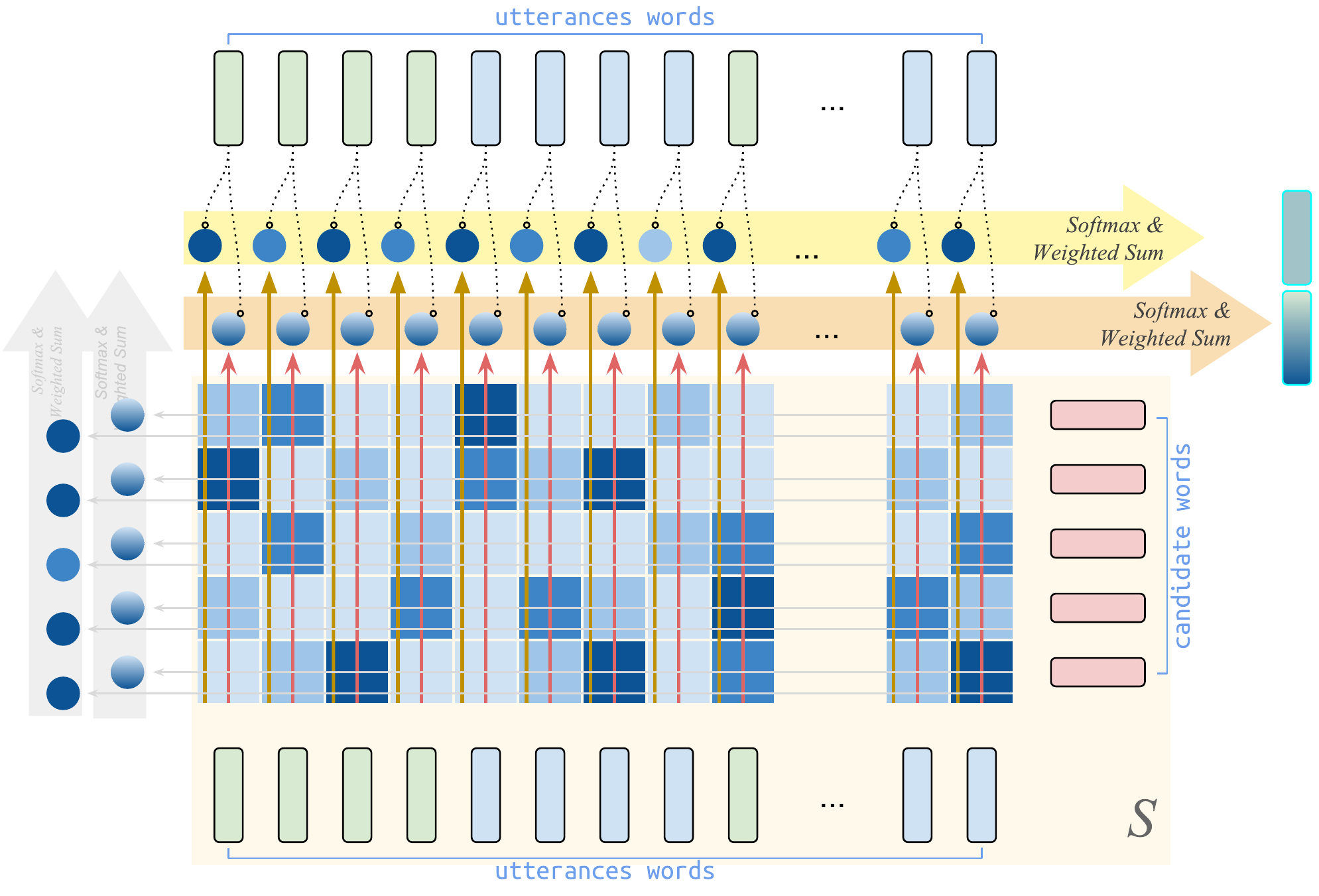}
  \caption{Illustrator of inter-attention with mean and max pooling. The brown and violet color arrows are column-wise max and mean operation respectively. The output vectors are the summation of utterances word embeddings weighted by the mean and max values. The dotted lines at the top of the figure denote that the word embeddings (green and blue rounded rectangles) are weighted by the values (blue circles). Weighted summation of the candidate word vectors (pink round corner rectangles) is omitted for simplicity here.}
  \label{fig:mcan-mean-max-pooling}
\end{figure}

\begin{figure}[t!]
  \centering
  \includegraphics[width=1\linewidth]{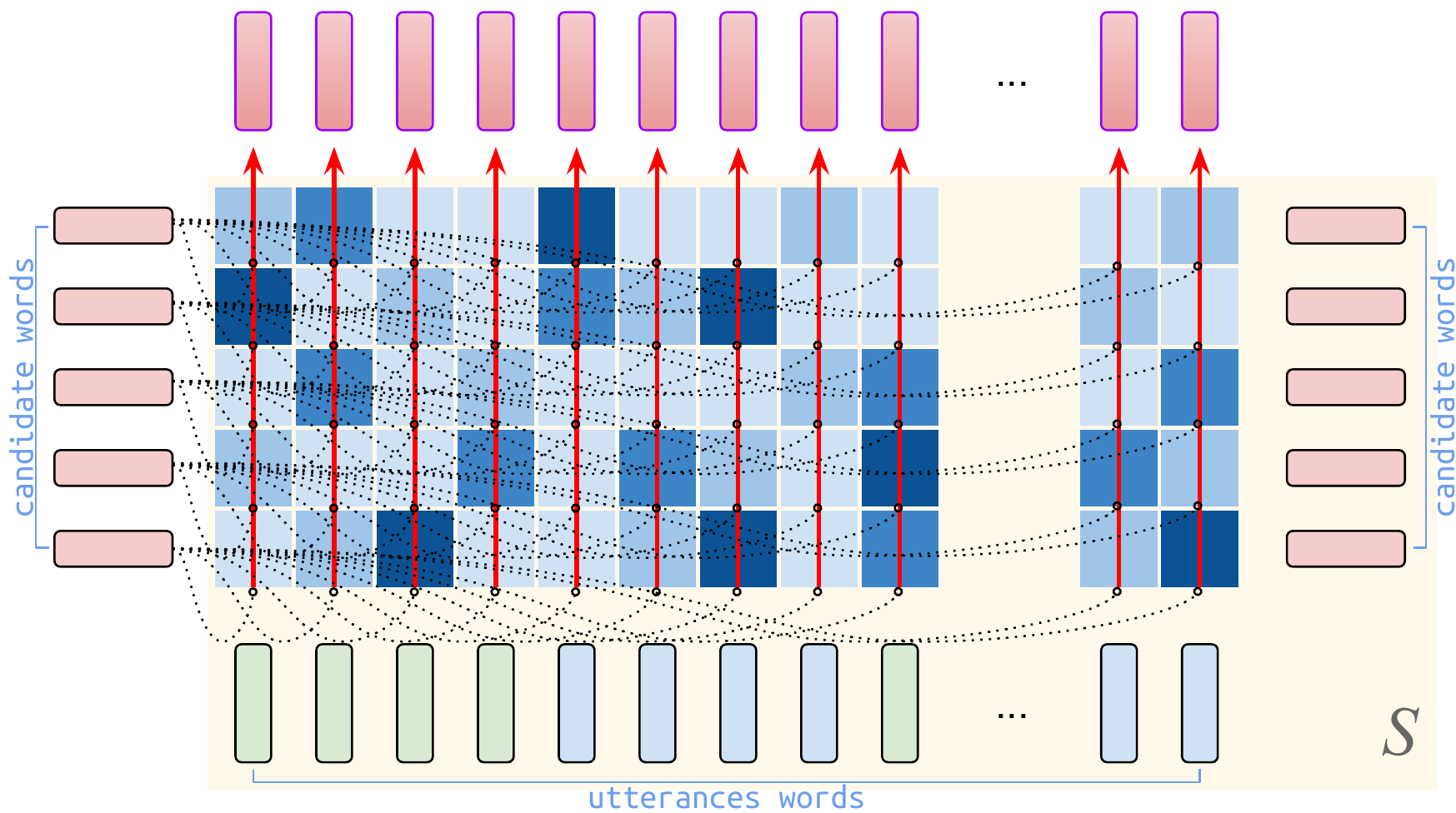}
  \caption{Illustration of inter-attention with alignment pooling. The red arrows implies summation of the candidates (pink round corner rectangles) weighted by the value of the column.}
  \label{fig:mcan-align-pooling}
\end{figure}

Before applying any attention operation, every word will by be transformed with the highway layer, the attention mechanisms include:

\paragraph{Intra-attention} for a sequence $d$, a similarity matrix $S$ is calculated as
  \begin{align}
   s_{i,j} = {w'}_i^{T} M {w'}_j, \nonumber
  \end{align}
  where $w_i$ and $w_j$ are $i$-th and $j$-th word of $d$ respectively, and $M$ is a parameter to learn.
  For $w_j$, the attention are then used to weighted sum over contexts to form a new representation: 
    \begin{align}
    & w'_{intra} = \sum_{i} \frac{\exp(s_{i,j}) }{\sum_{k} \exp ({s_{i, k}})} w'_i, \nonumber
  \end{align}
  where the weights are computed by performing softmax operation over columns of the similarity matrix $S$, and the intra-attention results for $q$ are calculated similarly.
\paragraph{Inter-attention} a similarity matrix $S$ is calculated as
  \begin{align}
    s_{i,j} = {w'}_{d,i}^T M {w'}_{q,j}, \nonumber
  \end{align}
   where $w'_{d,i}$ and $w'_{q,j}$ are $i$-th and $j$-th word of $d$ and $q$ respectively.
  The attention results are calculated differently for different pooling mechanisms ($M$ is not shared across different pooling mechanisms), as illustrated in Figure\ref{fig:mcan-mean-max-pooling} and Figure\ref{fig:mcan-align-pooling}:
  \begin{itemize}
  \item \emph{max-pooling}: the attention results are calculated as
    \[  w'_{max} = 
        \begin{cases}
        \mathrm{softmax}(\max_{col}(S))^T q & \quad \text{if } w \in q \\
        \mathrm{softmax}(\max_{row}(S)) d  & \quad \text{if } w \in d 
        \end{cases}
    \]
  \item \emph{mean-pooling}: the attention results are calculated as
      \[  w'_{mean} = 
        \begin{cases}
        \mathrm{softmax}(\mathrm{mean}_{col}(S))^T q & \quad \text{if } w \in q \\
        \mathrm{softmax}(\mathrm{mean}_{row}(S))  d  & \quad \text{if } w \in d 
        \end{cases}
    \]
  \item \emph{alignment-pooling}: the attention results are calculated as
    \[  w'_{align} = 
        \begin{cases}
        \sum_{i} \frac{\exp (s_{i,j}) }{\sum_{k} \exp{(s_{i, k})}} w'_{d,i} & \quad \text{if } w = w_{q,j} \\
        \sum_{j} \frac{\exp (s_{i,j}) }{\sum_{k} \exp{(s_{k, j})}} w'_{q,j} & \quad \text{if } w = w_{d,i}
        \end{cases}
    \]
  \end{itemize}
  Finally, we can have a feature vector of twelve scalar features by interacting $w'$ with other features produced by attention approaches described above:
  \begin{align}
    f_{mcan}(w') = [
     & W_1 [w'; w'_{align}] ; && W_2 [w'_{align} - w']; \nonumber \\
     & W_3 [w'_{align} \odot w']; && W_4 [w'; w'_{intra}]; \nonumber \\ 
     & W_5 [w'_{intra} - w']; && W_6 [w'_{intra} \odot w']; \nonumber \\
     & W_7 [w'; w'_{mean}] ; && W_8 [w'_{mean} - w']; \nonumber \\
     & W_9 [w'_{mean} \odot w']; && W_{10} [w'; w'_{max}]; \nonumber \\
     & W_{11} [w'_{max} - w']; && W_{12} [w'_{max} \odot w']; \nonumber
       ],
       \label{eq:mcan_map}
  \end{align}
  where $W_i$ are compression matrices that map a vector into a scalar and are parameters to learn.

\subsection{Word Feature Augmentation}

Secondly, each word $w$ is augmented by concatenating the domain specific knowledge-grounded features and features extracted by MCAN after the word embeddings. The sequences of words with extra features in the dialogue contexts and the candidates are denoted as
\begin{align}
    & \tilde{w} = [w; F(w), f_{mcan}(w)], \nonumber
\end{align}
where $F(w)$ is a vector about knowledge-grounded features as specified in the task description section.

\subsection{Dynamic Pooling Recurrent Networks}
To encode contextual information, a dynamic pooling recurrent network is proposed, which contains two layers of recurrent units and one dynamic pooling layer between the two recurrent layers. 
In our model, a bi-directional LSTM is employed~\cite{hochreiter1997long,schuster1997bidirectional}, which focuses on encoding the utterance-level information as hierarchical recurrent neural networks (HRED) does~\cite{serban2016building}. 
The $i$th sequence ($i = 1, 2, \cdots, l$) is encoded as

\begin{align}
    & \overrightarrow{h}^1_{i, t}, \overrightarrow{c}^1_{i, t} = \overrightarrow{\mathrm{LSTM}}^1(\tilde{w}_{i, t - 1}^U, \overrightarrow{h}^1_{i, t - 1}, \overrightarrow{c}^1_{i, t}), \nonumber \\
    & \overleftarrow{h}^1_{i, t}, \overleftarrow{c}^1_{i, t} = \overleftarrow{\mathrm{LSTM}}^1(\tilde{w}_{i, t + 1}^U, \overleftarrow{h}^1_{i, t + 1}, \overleftarrow{c}^1_{i, t}), \nonumber
\end{align}
where $h$ and $c$ are the hidden states and cell states respectively.
Different from HRED, which encodes each utterance separately, the first layer of Dynamic Pooling LSTM Networks encodes utterances by concatenating the utterances as a single sequence.
Therefore, the initial LSTM hidden state of an utterance is the last hidden state from the encoded previous sequence:
\begin{align}
    & \overrightarrow{h}^1_{i, 0}, \overrightarrow{c}^1_{i, t} = 
    \overrightarrow{h}^1_{i - 1, n_i}, \overrightarrow{c}^1_{i - 1, n_i} \nonumber \\
    & \overleftarrow{h}^1_{i, 0}, \overleftarrow{c}^1_{i, t} = 
    \overleftarrow{h}^1_{i + 1, n_i}, \overleftarrow{c}^1_{i + 1, n_i} \nonumber
\end{align}

In the second part of the dynamic pooling recurrent network, the dynamic pooling layer is used to generate one vector representation $\hat{h}^1_i$ for each utterance $u_i$ by pooling dynamically based on the utterance length $n_i$ over the encoded hidden states from the first bidirectional recurrent layer:
\begin{align}
  \hat{h}^1_i = \max([\overrightarrow{h}_{i,1}; \overleftarrow{h}_{i,1}], [\overrightarrow{h}_{i, 2},  \overleftarrow{h}_{i, 2}], \cdots, [\overrightarrow{h}_{i, n_i}; \overleftarrow{h}_{i, n_i}]), \nonumber
\end{align}
where $\max(.)$ is the operation of max pooling of the vectors over dimensions.
Finally, there is another bi-directional LSTM layer, which encodes utterance-level representations
\begin{align}
  & \overrightarrow{h}^2_i, \overrightarrow{c}^2_i = \overrightarrow{\mathrm{LSTM}}^2(\hat{h}^1_i, h^2_{i-1}, c^2_{i-1}), \nonumber \\
  & \overleftarrow{h}^2_i, \overleftarrow{c}^2_i = \overleftarrow{\mathrm{LSTM}}^2(\hat{h}^1_i, h^2_{i+1}, c^2_{i+1}), \nonumber
\end{align}
and the last LSTM cell state is used as the dialogue-level representation:
\begin{align}
  r^C = [\overrightarrow{c}^2_l; \overleftarrow{c}^2_1]. \nonumber
\end{align}

\subsection{Candidate Selection}
Each candidate $x_j$ is encoded by the first LSTM layer in the Dynamic Pooling LSTM:
\begin{align}
    & \overrightarrow{h}^x_{j, t}, \overrightarrow{c}^x_{j, t} = \overrightarrow{\mathrm{LSTM}}^1(\tilde{w}_{j, t - 1}^U, \overrightarrow{h}^x_{j, t - 1}, \overrightarrow{c}^x_{j, t}), \nonumber \\
    & \overleftarrow{h}^x_{j, t}, \overleftarrow{c}^x_{j, t} = \overleftarrow{\mathrm{LSTM}}^1(\tilde{w}_{j, t + 1}^U, \overleftarrow{h}^x_{j, t + 1}, \overleftarrow{c}^x_{j, t}), \nonumber
\end{align}
and max pooling is applied over the outputs to get the representation of the candidate:
\begin{align}
  r^X_j = \max([\overrightarrow{c}^x_{j, 1}; \overleftarrow{c}^x_{j, 1}],
  [\overrightarrow{c}^x_{j, 2}; \overleftarrow{c}^x_{j, 2}],
  \cdots,
  [\overrightarrow{c}^x_{j, m_jj}; \overleftarrow{c}^x_{j, m_j}]). \nonumber
\end{align}
Then the probability of the candidate $x_j$ being the correct response $y_j$ is calculated as
\begin{align}
  p(x_j) = \sigma(H_1(H_2([r^C; r^X_j; r^C \odot r^X_j; r^C - r^X_j]))), \nonumber
\end{align}
where $H_1, H_2$ are highway layers \cite{srivastava2015highway} with ReLU activation.
The binary cross entropy function is utilized as the objective:
\begin{align}
  L(U, X, Y) = \sum_{j=1}^k y_j \log p(x_j) + (1 - y_j) \log (1 - p(x_j)). \nonumber
\end{align}

\section{Experiments}

To evaluate the performance of the proposed RAP-Net, we conduct experiments on the two datasets provided by DSTC7-Track1, and compare our results with two baseline systems.


\subsection{Dataset}

DSTC7-Track1 contains two goal-oriented dialogue datasets --- 1) Ubuntu data and 2) Advising data.
There are five subtasks in this challenge, where this paper focuses on the subtask 1, 3 and 4, because the same model architecture can be applied to these subtasks.
Here we briefly describe the settings for each subtask:
\begin{itemize}
  \item Subtask 1: There are 100 response candidates for each dialogue, and only one is correct.
  \item Subtask 3: There are 100 response candidates for each dialogue. The number of correct responses is between 1 and 5. Multiple correct responses are generated using paraphrases.
  \item Subtask 4: There are 100 response candidates for each dialogue, and an additional candidate \textit{no answer} should also be considered. The number of correct responses is either 0 or 1.
\end{itemize}

\subsection{Baseline Systems}

\begin{itemize}
  \item Dual Encoder \cite{lowe2015ubuntu}: uses two LSTMs with tied weights to encode the context $d = \{ u_1, u_2,\cdots,u_l \}$ and the response $x$ into fixed-length representations $c, r$, respectively.
  The final hidden state of LSTM is used to represent an input word sequence. 
  The probability of $x$ being the next utterance of $c$ is then calculated as 
  \begin{align}
    p = \sigma (c^T M r + b), \nonumber
  \end{align}
  where the matrix $M$ and bias $b$ are learned parameters.
  
  \item HRED~\cite{serban2016building}: has a similar structure as the Dual Encoder, but uses two LSTMs to encode context hierarchically. 
  Each utterance in the dialogue context is encoded separately by an utterance-level LSTM $LSTM^{1}$. 
  The encoded representations are then fed into a conversation-level LSTM $LSTM^{2}$ to produce the context representation $c$. A response $x$ is encoded by $LSTM^{1}$ into response representation $r$. The prediction is calculated similarly as the Dual Encoder above.
\end{itemize}

\subsection{Experimental Details}

We use pre-trained 300-dimensional word embeddings via \texttt{fasttext} \cite{mikolov2018advances} to initialize the embedding matrix and fix it during training.
The word embeddings of out-of-vocabulary (OOV) are initialized randomly.
In the advising dataset, the suggested courses $C_{suggested}$ and the prior courses $C_{prior}$ of the student are given along with a conversation.
Therefore, to explicitly utilize this knowledge in our model, we extract two features for each word and then concatenate them as the knowledge-grounded features, $F(w)$:
\begin{align}
  & F_1(w) =
    \begin{cases}
      1 & \quad \text{if } w \in C_{suggested} \\
      0 & \quad \text{otherwise.}  \\
    \end{cases} \nonumber \\
  & F_2(w) = 
    \begin{cases}
      1 & \quad \text{if } w \in C_{prior} \\
      0 & \quad \text{otherwise.} \\
    \end{cases} \nonumber\\
  & F(w) = [F_1(w); F_2(w)] \nonumber
\end{align}
Note that there is no additional knowledge provided for the Ubuntu dataset except for the subtask 5.
Therefore, only $f_{mcan}(w)$ is added to the word representations.

We use \texttt{adam} as our optimizer to minimize the training loss~\cite{kingma2014adam}.
The hidden layer size of LSTM is 128. The initial learning rate varies from 0.001 to 0.0001, which is a hyperparameter for tuning.
We train our models for 10 epochs and select the best-performing model based on the development set.

Following the official evaluation metrics, we use \emph{recall at 10} (R@10) and \emph{mean reciprocal rank} (MRR) to report the performance of our models.
The final score is the average of two metrics.

\begin{table*}[t!]
    \centering
    \begin{tabular}{ll|ccc|ccc}
    \toprule
    && \multicolumn{3}{c|}{Ubuntu} & \multicolumn{3}{c}{Advising} \\
    && R@10      & MRR        & Average        & R@10       & MRR        & Average \\ \midrule
    Baseline & (a) Dual Encoder                    & 62.5          & 36.23     & 49.39         & 25.8           & 11.81     & 18.81   \\
     & (b) HRED                    & 65.2          & 37.87     & 51.56         & 39.2           & 18.68     & 28.94   \\ \midrule
    RAP-Net & (c) DP-LSTM                    & 66.3          & 41.26     & 53.81         & 49.0           & 21.99     & 35.49   \\
    & (d) DP-LSTM+$f_{mcan}$         & \bf 76.7          & \bf 56.18     & \bf 66.45         & 51.0           & 25.80     & 38.40  \\
    & (e) DP-LSTM+$F(w)$             & -              & -          & -              & 72.9           & 38.07     & 55.50  \\
    & (f) DP-LSTM+$f_{mcan}$+$F(w)$  & -              & -          & -              & \bf 76.6 & \bf 42.84  & \bf 59.72  \\
    \bottomrule
    \end{tabular}
    \caption{Results on subtask 1 development sets (\%).}
    \label{tab:feature}
\end{table*}

\begin{table*}[t!]
    \centering
    \begin{tabular}{l|cccc|ccc|cccc}
    \toprule
    \multirow{2}{*}{Task} & \multicolumn{4}{|c}{Ubuntu}   & \multicolumn{3}{|c}{Advising Case 1} & \multicolumn{4}{|c}{Advising Case 2} \\
                         & R@10 & MRR    & Avg & Rank & R@10 & MRR    & Avg & R@10    & MRR      & Avg  & Rank \\
    \midrule
    Subtask 1            & 81      & 64.86 & 72.93 & 3  & 80.4     & 49.14 & 64.77  & 61         & 30.61   & 45.81  & 2 \\
    Subtask 3            & -         & -      & -      & - & 68.4     & 39.34 & 53.87  & 60.4        & 31.71   & 46.05  & 3  \\
    Subtask 4            & 84.1     & 63.17 & 73.63 & 2 & 84.2     & 45.31 & 64.75  & 64.0        & 30.70   & 47.35 & 3\\
    \bottomrule
    \end{tabular}
    \caption{The official testing results of our submitted systems. Two different test sets for advising are provided. Note that in subtask 3, only advising dataset is provided for training and evaluating. The rankings of this challenge are based on the average score.}
    \label{tab:official}
\end{table*}

\subsection{Results}
To explicitly validate the effectiveness of the proposed model and auxiliary features, we compare the performance between our model and the baseline systems.
Table~\ref{tab:feature} shows the empirical results on the development set of the subtask 1.

\paragraph{Dynamic pooling LSTM}
Our dynamic pooling LSTM outperforms HRED in terms of all metrics on both datasets, especially on the advising dataset.
The results show that concatenating utterances into a single sequence can benefit conversation encoding.

\paragraph{MCAN feature}
Adding $f_{mcan}(w)$ as an auxiliary feature (row (d)) further improves the performance by a large margin on the ubuntu dataset, yielding a 23.5\% relative improvement.
It demonstrates that the MCAN feature $f_{mcan}(w)$ also helps our model achieve slightly better results on the advising dataset.

\paragraph{Knowledge-grounded feature}
For advising dataset, we extract a 2-dimensional knowledge-grounded feature $F(w)$ to enhance word representations. 
As shown in Table~\ref{tab:feature}, adding $F(w)$ (row (e)) yields a 56.4\% relative improvement, which is significantly greater than the improvement of adding $f_{mcan}(w)$.
The results show the difficulty of solving this task on the advising dataset without any prior knowledge.
The effectiveness of our knowledge-grounded feature $F(w)$ shows that identifying course names is crucial for this dataset. 
The best model on the advising dataset is the dynamic pooling LSTM with both auxiliary features added to the input word representations (row (f)), achieving about 66\% and 60\% average performance for ubuntu and advising datasets respectively.

\subsection{Official Evaluation}

In the DSTC7 challenge, the proposed systems are submitted for official evalution.
For each subtask, our submitted system consists of several models with different hyperparameters and auxiliary features.
Using different features gives our model multiple perspectives to the data and hence improves the prediction accuracy.
The official evaluation results are shown in Table \ref{tab:official}, and 
Table \ref{tab:rank} shows the final rankings of our results for all subtasks.
Note that the model for the subtask 5 is the single model, which performs worse than the ensemble one as the subtask 1.
In the official evaluation, the superior performance and the achieved rankings across different subtasks significantly demonstrate the effectiveness of the proposed model.
Considering that our rankings are either 2 or 3 among 20 teams, we argue that the proposed RAP-Net can successfully estimate the relatedness between dialogues and responses and generalize across different datasets.

\begin{table}[t!]
    \centering
    \begin{tabular}{lcc}
    \toprule
              & Ubuntu & Advising \\ \midrule
    Subtask 1 & 3      & 2        \\ 
    Subtask 3 & -      & 3        \\ 
    Subtask 4 & 2      & 3        \\ 
    Subtask 5 & 6      & 4        \\ 
    \midrule
    Overall   & 4      & 3        \\ \bottomrule
    \end{tabular}
    \caption{Official rankings of our systems on each subtask. The overall ranking considers all subtasks.}
    \label{tab:rank}
\end{table}

\begin{table}[t!]
    \centering
    \begin{tabular}{l|ccc}
    \toprule
    \multicolumn{1}{l|}{Model}                      & R@10 & MRR    & Average \\
    \midrule
    \multicolumn{1}{l|}{DP-LSTM + $f_{mcan}$}  & 76.7     & 56.18 & 66.45  \\
    ~~- inter-attention                              & 69.4     & 43.92 & 56.66  \\
    ~~- intra-attention                           & 76.3     & 55.86 & 66.08  \\
    ~~- highway encoder                           & 75.8     & 54.94 & 65.37  \\
    ~~- dynamic pooling                           & 76.1     & 55.18 & 65.67  \\ \bottomrule
    \end{tabular}
    \caption{Ablation results on Ubuntu development set (\%).}
    \label{tab:ablation}
\end{table}

\begin{figure*}[t!]
    \centering
    \subfloat[Context]{\includegraphics[width=0.9\textwidth]{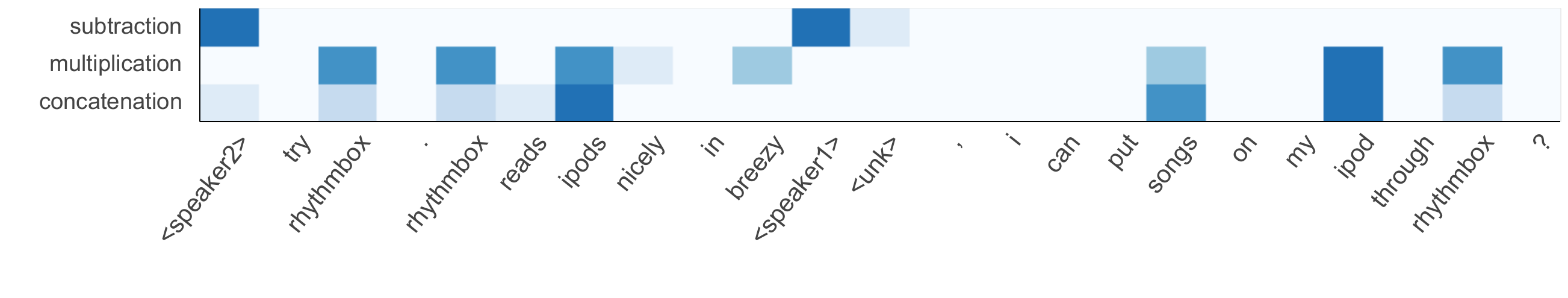}} \\
    \subfloat[Response]{\includegraphics[width=0.55\textwidth]{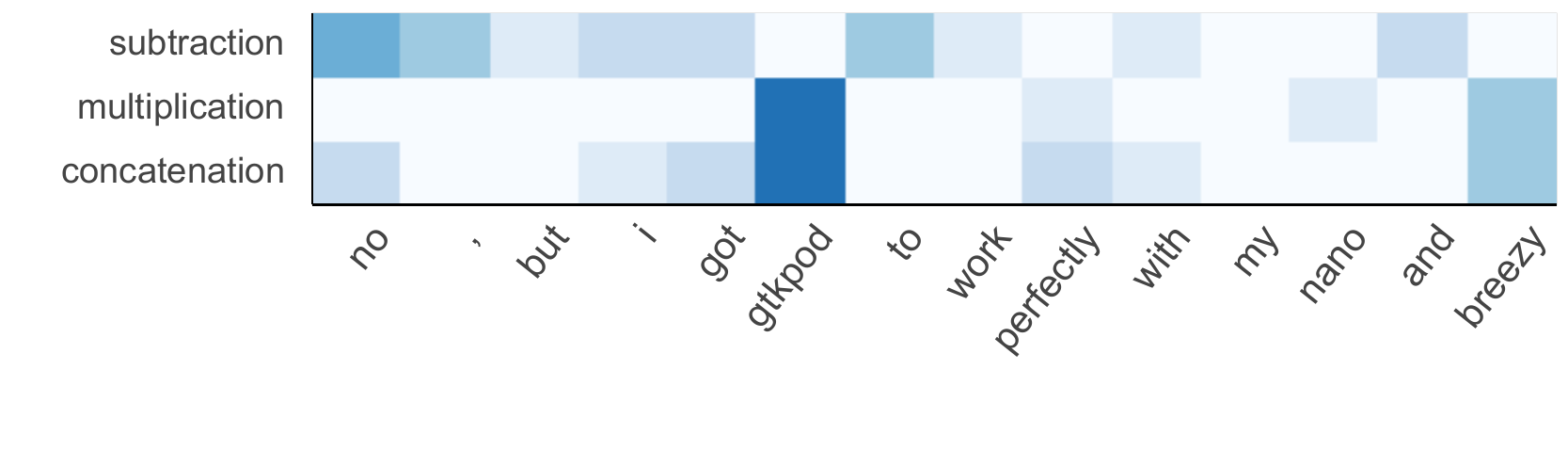}}
    \caption{Visualization of attention scores. We plot the attention scores $f_{mcan}$ of two sequences from Ubuntu development set: (a) A partial conversation and (b) The correct response corresponding to the conversation. The conversation is truncated to the last two sentences due to width limitation. Darker color represents higher attention score. Note that each row is normalized separately since the range of values varies for each dimension.}
    \label{fig:attention-score}
\end{figure*}

\subsection{Ablation Study}

To further understand the contribution of each component, we conduct an ablation test on the RAP-Net model.
Table \ref{tab:ablation} shows the ablation results on the ubuntu subtask 1 development set.
We remove one component at each time and evaluate the resulting model using R@10 and MRR. Note that after removing dynamic pooling, the last hidden state of an LSTM is used as the sequence-level representation.
This setting is equivalent to HRED with an additional feature $f_{mcan}$.

The ablation results show that the inter-attention is the most crucial component to our model, because the average score drops drastically by almost 10 if it is removed, demonstrating the importance of modeling the interaction between the conversation and the response for this task.
It is found that removing each of them results in a reduction of roughly 1 in terms of the average score, so the highway encoder and the dynamic pooling also slightly contribute to the improvement.
Furthermore, the intra-attention benefits least to performance, which is similar to the findings in the prior work~\cite{tay2018multi}.

\subsection{Attention Analysis}

As described in the previous section, the attention $f_{mcan}$ is a key feature in our framework.
To deeply investigate this feature, we examine its numerical value to perform qualitative analysis.
An example of attention scores for each word in a sequence is shown in Figure \ref{fig:attention-score}. 
It can be found that the features extracted by mean pooling, max pooling and intra-attention are always equal or close to zero with no obvious pattern, so we only plot features extracted by alignment-pooling for simplicity.
The x-axis indicates the words in the context or response, and the y-axis represents different compression methods described in the MCAN section.

From Figure \ref{fig:attention-score}, we observe that the attention has the ability to model word overlapping between two sequences.
For example, the word \emph{breezy} appears in both sequences, and it has a relatively larger attention score.
In addition to the ability to model explicit word overlapping, MCAN can also identify words that are relevant to the other sequence.
Here MCAN gives \emph{rhythmbox} and \emph{ipod} larger scores than other words in the context, even though they do not appear in the response.
The reason is that words such as \emph{gtkpod} and \emph{breezy} in the response are related to \emph{ipod}, so the model correctly identifies the words that are relevant in the context.
Similarly, the word \emph{gtkpod} in the response obtains the largest attention score, because it is the most relevant to the context.

The features extracted by multiplication and concatenation shows similar patterns. However, the features extracted by subtraction seems to be only activated by \texttt{<speaker>} and \texttt{<unk>} tokens or other function words.
The probable reason is that this dimension assists the encoder to recognize unimportant words.
We should note that these observed patterns are not consistent over different runs. 
Generally there is at least one dimension that models word relevance across sequences, and at least one dimension that recognizes unimportant words. 

\section{Conclusion}
This paper proposes a novel framework, recurrent attention pooling networks (RAP-Net), which focuses on precisely measuring the relations between dialogue contexts and the responses for dialogue response selection.
The DSTC7 experiments are conducted to evaluate the proposed model, where multi-cast attention network (MCAN) and our proposed knowledge-grounded features are proved to be useful, and each attention and pooling mechanism is demonstrated to be effective.
In sum, RAP-Net is capable of capturing the salient information from dialogues and is good at selecting a proper response for two different types of dialogue data.
In the future, the proposed model can be evaluated on other retrieval-based tasks to test the model capability of generalization.

\bibliography{main}
\bibliographystyle{aaai}

\end{document}